\documentclass[10pt,twocolumn,letterpaper]{article}

\usepackage[pagenumbers]{cvpr} % To force page numbers, e.g., for an arXiv version

\definecolor{cvprblue}{rgb}{0.21,0.49,0.74} % Define a custom color for links
\colorlet{dark-green}{green!70!black} % Define a dark green color

\usepackage[pagebackref,breaklinks,colorlinks,allcolors=cvprblue]{hyperref} % Configure hyperref

\usepackage{graphicx}
\usepackage{amsmath}
\usepackage{amssymb}
\usepackage{multirow}   % For multi-row cells in tables
\usepackage{booktabs}   % For enhanced table formatting
\usepackage{wasysym}	% For additional symbols
\usepackage{microtype}  % For better typography
\usepackage{bm}		 % For bold math symbols
\usepackage{float}	  % For improved figure/table placement
\usepackage{colortbl}
\usepackage{algorithm} % Recommended for algorithms
\usepackage{algorithmic} % Recommended for algorithms
\usepackage{pifont}
\usepackage{marvosym}
\usepackage{soul}

\def\paperID{311}	  % *** Enter the Paper ID here
\def\confName{CVPR}	% Conference name
\def\confYear{2025}	% Conference year

\title{FocalCount: Towards Class-Count Imbalance in Class-Agnostic Counting}

\author{
Huilin Zhu$^{1,2}$, 
Jingling Yuan$^{1,\ast}$, 
Zhengwei Yang$^{3}$, 
Yu Guo$^{4,2}$, 
Xian Zhong$^{1,\ast}$, 
Shengfeng He$^{2}$ \\
$^{1}$ Hubei Key Laboratory of Transportation Internet of Things, Wuhan University of Technology \\
$^{2}$ School of Computing and Information Systems, Singapore Management University \\
$^{3}$ School of Computer Science, Wuhan University \\
$^{4}$ School of Navigation, Wuhan University of Technology \\
{\tt\small yjl@whut.edu.cn, zhongx@whut.edu.cn}
} 

\begin{document}
\maketitle

\begin{abstract}
In class-agnostic object counting, the goal is to estimate the total number of object instances in an image without distinguishing between specific categories. Existing methods often predict this count without considering class-specific outputs, leading to inaccuracies when such outputs are required. These inaccuracies stem from two key challenges: 
1) the prevalence of single-category images in datasets, which leads models to generalize specific categories as representative of all objects, and 
2) the use of mean squared error loss during training, which applies uniform penalization. This uniform penalty disregards errors in less frequent categories, particularly when these errors contribute minimally to the overall loss. 
To address these issues, we propose {FocalCount}, a novel approach that leverages diverse feature attributes to estimate the number of object categories in an image. This estimate serves as a weighted factor to correct class-count imbalances. Additionally, we introduce {Focal-MSE}, a new loss function that integrates binary cross-entropy to generate stronger error gradients, enhancing the model's sensitivity to errors in underrepresented categories. Our approach significantly improves the model's ability to distinguish between specific classes and general counts, demonstrating superior performance and scalability in both few-shot and zero-shot scenarios across three object counting datasets. The code will be released soon.

\end{abstract}

\section{Introduction}

% 背景
% 然而这些方法只能对固定的类用固定的训练集, 缺乏泛化性, 因此, 类不可知目标计数越来越受到人们的关注。

Object counting is a fundamental task in computer vision, involving the estimation of object quantities in images or videos. Traditionally, this task has focused on specific categories, such as crowds, vehicles, and cells~\cite{tyagi2023degpr, arteta2016counting, mundhenk2016large, babu2022completely}. However, these methods are limited by their reliance on predefined categories and specific training datasets, which restrict their generalizability. To address this, class-agnostic object counting, operating independently of predefined categories, has gained significant attention for its broader applicability.
Class-agnostic object counting methods can be categorized into three main types: visual-prompt-based, textual-prompt-based, and reference-less. Visual-prompt methods perform well using image references but require bounding box annotations during inference. Textual-prompt methods allow category specification without bounding boxes but struggle due to the semantic gap between text and images, making category association challenging. Reference-less methods minimize annotation requirements but do not permit category specification, often resulting in the counting of all visible objects, which may not align with specific objectives.

\begin{figure}
	\centering
	\includegraphics[width = \linewidth]{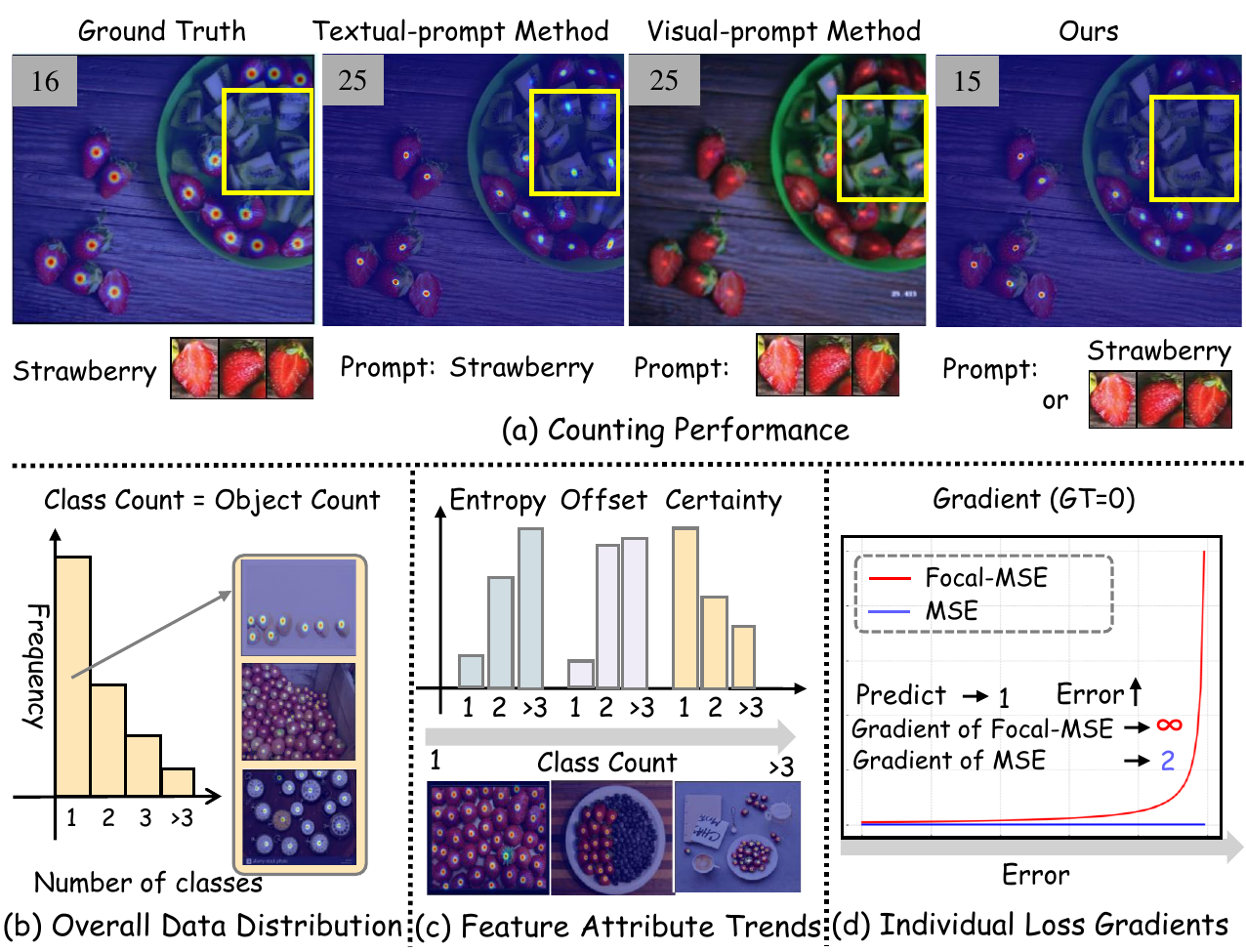}
	\caption{\textbf{Illustration of Object Counting.} 
	(a) Miscounting of unspecified classes by CounTR (few-shot visual prompt) and CLIP-Count (zero-shot textual prompt). 
	(b) Object count distribution showing single-category dominance in \textsc{FSC-147}. 
	(c) Impact of category diversity on counting accuracy using feature attributes. 
	(d) Gradient comparison between Focal-MSE and MSE.
}
	\label{fig:1}
\end{figure}

To enhance category-specific counting, we conduct an extensive analysis of textual-prompt and visual-prompt methods, revealing key insights. As shown in \cref{fig:1}(a), these methods often count all objects in an image, regardless of prompts, resembling the behavior of reference-less methods and undermining precise object counting. Further analysis indicates that model bias contributes significantly to this issue. As illustrated in \cref{fig:1}(b), many images in object counting datasets contain only one category, with the total object count matching the specified category. This imbalance causes models to indiscriminately count all objects, exacerbating the problem. Additionally, \cref{fig:1}(d) shows that MSE loss, the most commonly used loss function in object counting, penalizes large and small prediction errors equally. In datasets with few multi-category images, this loss function fails to address small counts of non-specified categories effectively.

To address indiscriminate object counting, two key objectives must be met: mitigating data imbalances by increasing focus on multi-category images, and implementing a loss function that enhances sensitivity to errors in regions with non-specified classes, thereby improving supervision. A common strategy for handling data imbalance is reweighting~\cite{regmi2024reweightood, he2024gradient, qiu2023simple}, which assigns higher weights to minority samples. However, in class-agnostic counting, each image is labeled with only one category, making it difficult to infer the number of categories based solely on class labels. Despite this, the number of categories within an image can be inferred. As shown in \cref{fig:1}(c), there is a correlation between discernible patterns in feature attributes and the number of categories in images.

\begin{figure}
	\centering
	\includegraphics[width = \linewidth]{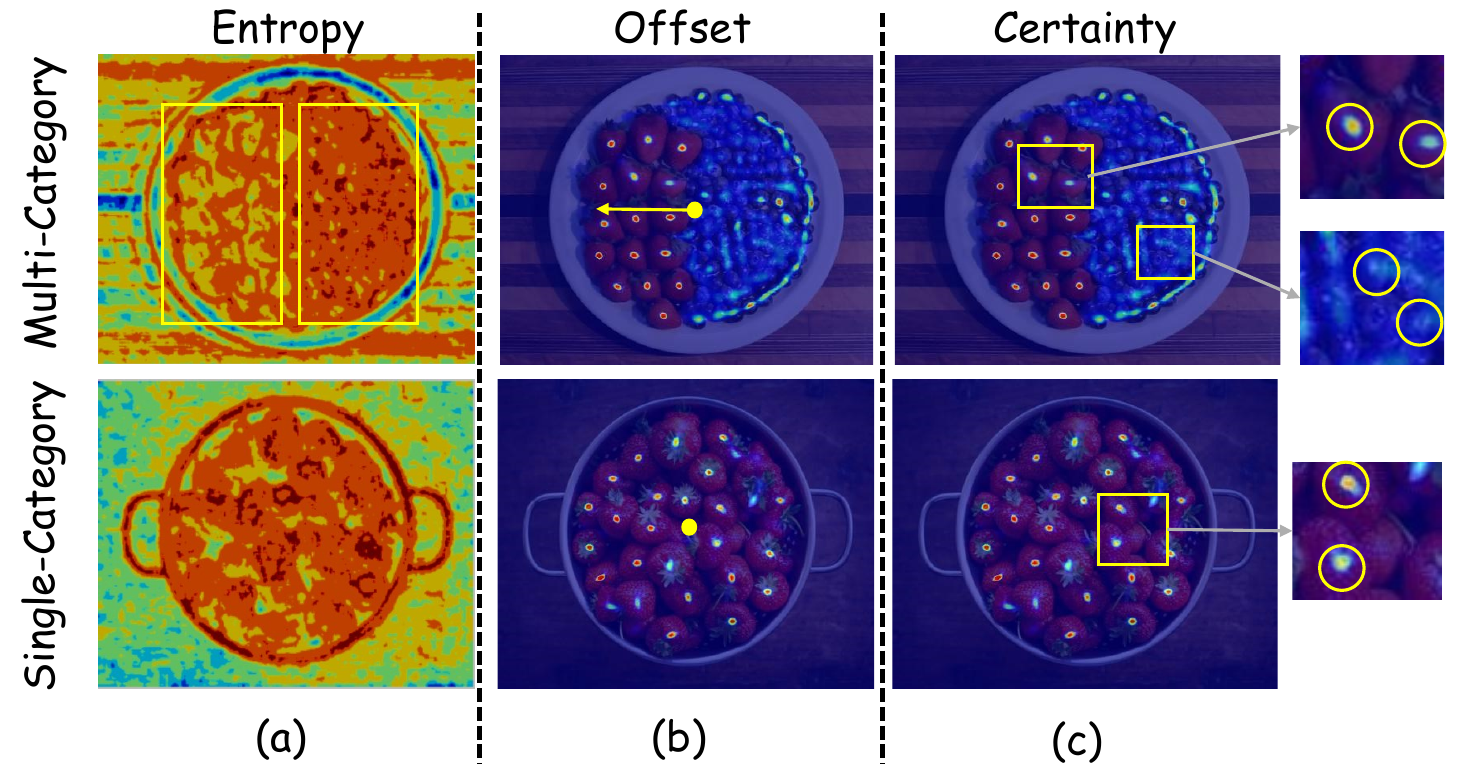}
	\caption{\textbf{Illustration of Feature Attributes in Single-Category versus Multi-Category Images.}}
	\label{fig:4}
\end{figure}

Moreover, as shown in \cref{fig:4}, distinct differences in feature attributes are observed between multi-category and single-category images. Specifically, \cref{fig:4}(a) shows that images with multiple categories exhibit higher entropy, reflecting greater information complexity. In contrast, single-category images typically display a more centralized specified category, while multi-category images show an offset in the specified category area, as depicted in \cref{fig:4}(b). Interestingly, although objects from non-specified categories are counted, their certainty is significantly lower than that of specified categories, as shown in \cref{fig:4}(c).

These insights led to the development of FocalCount, a method that assigns different weights to data based on the likelihood of containing multiple categories. Additionally, as shown in \cref{fig:1}(d), while Focal-MSE provides more adaptive gradients for reducing errors in localized regions compared to MSE, it alone lacks the precision required for exact quantity supervision. To address this, we integrate Focal-MSE into a curriculum learning framework, where it is used in the early stages of training alongside error-sensitive entropy. This approach improves precise quantity supervision and enhances the model's sensitivity to errors in non-specified categories.

In summary, our contributions are fourfold:

\begin{itemize}
	\item We are the first to identify that the prevalent data imbalance, where single-category images dominate in object counting datasets, leads most methods to count all objects in an image rather than focusing on the specified category.
 
	\item We propose FocalCount, a novel method that addresses data imbalance by estimating image category counts through the relation between feature attributes and the number of categories.
 
	\item We introduce Focal-MSE loss, a new loss function that integrates binary cross-entropy to modulate training with stronger error gradients, enhancing the model's sensitivity to errors in non-specified categories.
 
	\item Extensive experiments on three object counting datasets in both few-shot and zero-shot settings demonstrate the state-of-the-art accuracy and generalizability of FocalCount.

\end{itemize}

\begin{figure*}
	\centering
	\includegraphics[width = 0.98\linewidth]{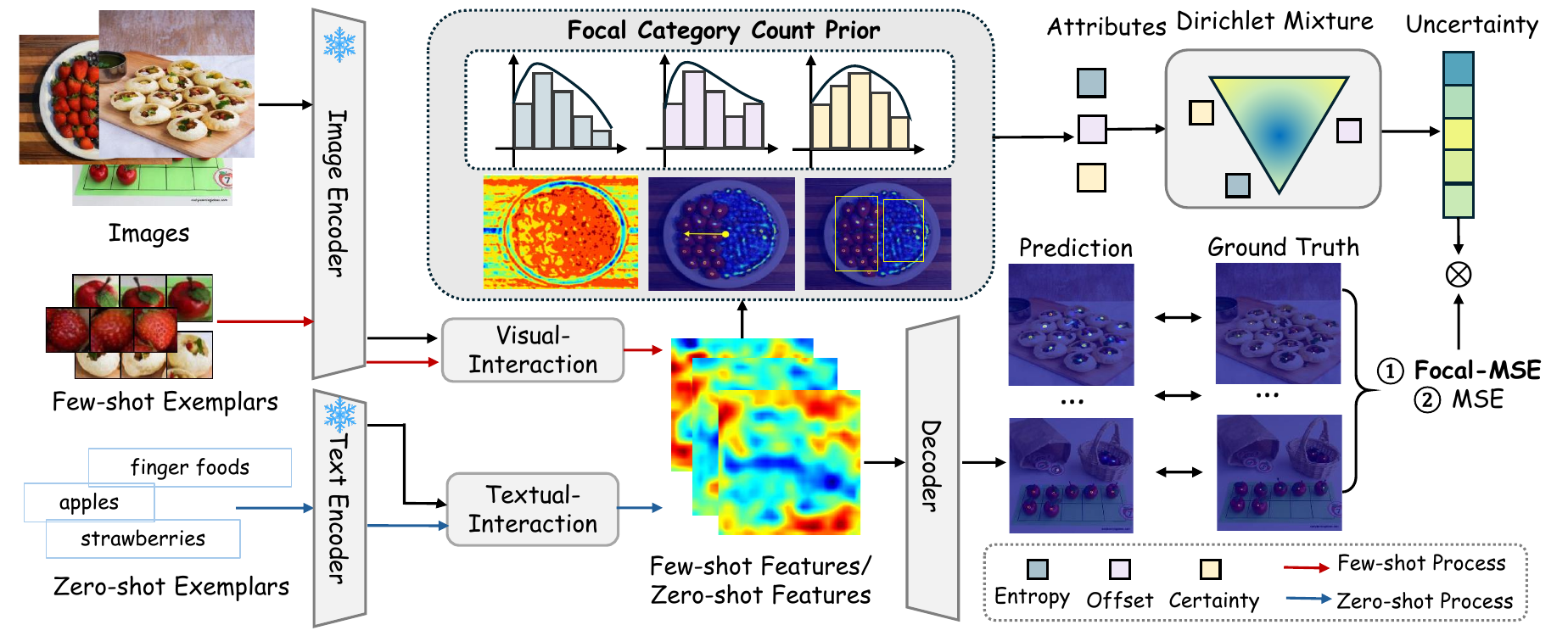}
	\caption{\textbf{Overview of the FocalCount Method.} 
	(1) Mitigating data imbalance by estimating the Focal Category Count Prior using entropy, offset, and feature certainty.
	(2) Employing dual-phase curriculum learning: initially applying Focal-MSE to enhance quantity supervision and sensitivity to errors in unspecified categories, followed by MSE for precise learning.}
	\label{fig:2}
\end{figure*}

\section{Related Work} 

\subsection{Object Counting}
Object counting plays a crucial role in applications such as public safety, administration, and labor efficiency. Traditional methods~\cite{ranjan2022exemplar, shi2022represent, yang2021class, you2023few} are restricted to fixed categories and require retraining when new categories are introduced. In contrast, class-agnostic counting~\cite{lu2019class, gong2022class, nguyen2022few, lin2024gramformer, du2023domain} offers more flexible solutions, supporting scenarios with limited data and enabling few-shot, reference-less, and zero-shot counting methods.

\vspace{-10pt}
\paragraph{Few-shot object counting} addresses scenarios with limited annotated data. GMN~\cite{lu2019class} formulates class-agnostic counting as a matching task, leading to FamNet~\cite{ranjan2021learning}, which incorporates ROI Pooling. BMNet~\cite{shi2022represent} introduces a bilinear matching network to refine similarity assessments. LOCA~\cite{djukic2023low} enhances feature representation and exemplar adaptation, while CounTR~\cite{liu2022countr} uses transformers to scale counting tasks. CACViT~\cite{WangX0024} integrates Vision Transformers (ViT) into object counting for improved performance.

\vspace{-10pt}
\paragraph{Zero-shot object counting} operates without the need for category-specific training data. CLIP-Count~\cite{jiang2023clip} leverages CLIP to encode text and images separately, enabling semantic associations, while VLCount~\cite{kang2023vlcounter} enhances text-image association learning. PseCo~\cite{huang2023point} introduces a SAM-based framework for segmentation, dot mapping, and detection, offering broad applicability but requiring significant computational resources.

\vspace{-10pt}
\paragraph{Reference-less object counting} do not rely on specific references. ZSC~\cite{xu2023zero} generates prototypes using textual inputs and filters image patches, reducing labeling requirements but facing scalability challenges. While these methods are flexible and scalable, they often struggle with accuracy across diverse categories due to the absence of multi-category density map labels in most datasets.

\subsection{Counting Loss}
The most commonly used loss function in object counting is mean squared error (MSE) loss~\cite{ranjan2021learning}, which effectively measures the difference between predicted and ground truth maps. To capture associations, GMN~\cite{lu2019class} introduces a similarity loss that quantifies the relations between predicted and actual similarities, while FamNet~\cite{ranjan2021learning} incorporates a perturbation loss to increase robustness. LOCA~\cite{djukic2023low} adds an auxiliary loss to support multi-channel learning. Other methods~\cite{jiang2023clip, huang2023point} employ an InfoNCE-based contrastive loss~\cite{oord2018representation} to distinguish target regions from the background, while VA-Count~\cite{zhu2024zero} uses contrastive loss to differentiate between known and unknown classes.

However, these loss functions do not address the imbalance between single-label and multi-label data, which often leads models to indiscriminately count all objects. Inspired by Focal Loss~\cite{lin2017focal}, we propose Focal-MSE, an error-sensitive loss function that enhances sensitivity to specific regions. Focal-MSE promotes intra-class compactness and inter-class distinctiveness, transforming counting into a probability metric for specified categories and ensuring precise counting while overcoming the limitations of traditional loss.

% \subsection{Dirichlet Distribution}

\section{Proposed Method}
\subsection{Problem Definition}
Class-agnostic object counting models often suffer from a bias towards single-category images due to dataset imbalances, resulting in inaccurate object counts that fail to focus on specified categories. We propose {FocalCount}, a novel method to address this data imbalance and improve model sensitivity to errors. As illustrated in \cref{fig:2}, {FocalCount} first analyzes feature attribute trends to emphasize images with multiple categories (See \cref{sec 3.1}). Then, we apply {Focal-MSE}, an error-sensitive loss function, to improve accuracy in detecting density map errors, particularly for unspecified categories in multi-category images (See \cref{sec 3.2}).

Let the training set of images be denoted as $\{I_i\}_{i=0}^n$, and the exemplars (either visual or textual prompts) as $\{P_i\}_{i=0}^n$. The output features are represented as $\{F_i\}_{i=0}^n$, the predicted density maps as $\{M^p_i\}_{i=0}^n$, and the ground truth density maps as $\{M^g_i\}_{i=0}^n$. The objective is to minimize the following loss function, where $U_i$ denotes data adjustment weights, $\mathcal{L}$ is the loss function, and $n$ is the number of images:
\begin{equation}
	\mathrm{Objective} = \min \sum_{i=0}^n U_i \mathcal{L} \left( M^p_i, M^g_i \right).
\end{equation}

% In this paper, the training set of original images is defined as $\{I_i\}_{i = 0}^n$, and the exemplars, which can be image or text prompts, as $\{P_i\}_{i = 0}^n$. The features output by the image encoder are $\{F_i\}_{i = 0}^n$, the predicted density maps are $\{M^p_i\}_{i = 0}^n$, and the ground truth density maps are $\{M^g_i\}_{i = 0}^n$. 
% Our goal is to minimize by adjusting data importance through weights:
% By adjusting the data importance through weights, our objective is to minimize:
% \begin{equation}
% 	\mathrm{Objective} = \min \sum_{i = 0}^n w_i \mathcal{L}(M^p_i, M^g_i), 
% \end{equation}
% where $W = \{w_i\}_{i = 0}^n$ represents the data adjustment weights and $\mathcal{L}$ is the loss function. $n$ represents the number of images in the dataset.

\begin{figure*}
	\centering
	\includegraphics[width = 0.98\linewidth]{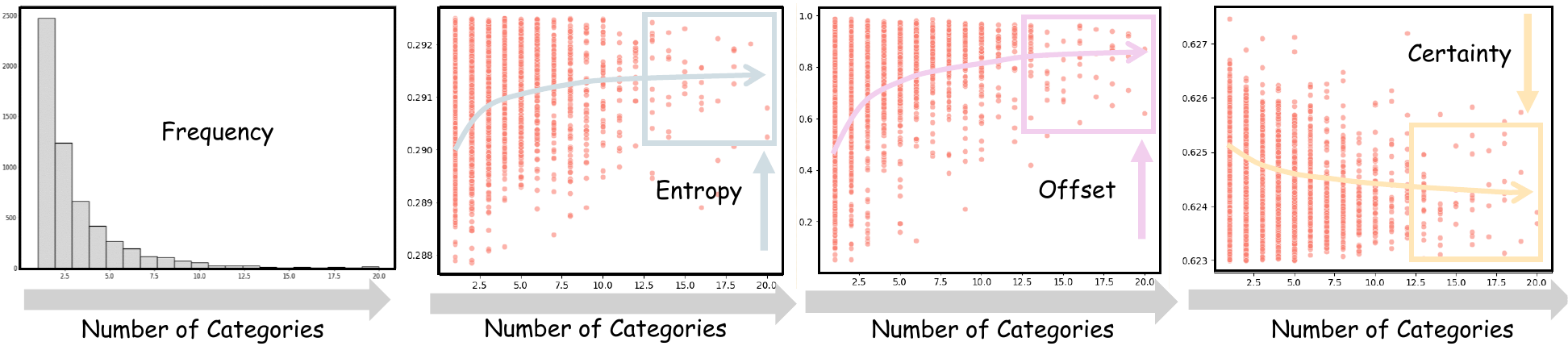}
	\caption{\textbf{Relation between Number of Categories and Attributes in \textsc{Pascal VOC}~\cite{everingham2010pascal}.} As the number of categories increases, entropy and offset rise, while certainty declines, indicating greater complexity.}
	\label{fig:3}
\end{figure*}

\subsection{Focal Imbalance Rectification}
\label{sec 3.1}

\subsubsection{Focal Category Count Prior.}
In object counting datasets, images often contain only a single category, making it challenging to estimate the number of categories from labels alone. However, the presence of multiple categories can leave identifiable traces. By analyzing feature attributes, we can infer the likelihood of multiple categories. \cref{fig:3} shows the relation between category count and three attributes: entropy and offset, which are positively correlated with category count, and certainty, which is negatively correlated. Combining these attributes improves the reliability of category count estimation.

% \vspace{-10pt}
\subsubsection{Feature Extraction.}
Given a batch of images $\{I_i\}_{i=1}^n$ and exemplars $\{P_i\}_{i=1}^n$, we input them into the encoder $\phi(\cdot)$ to obtain features $\{F_i\}_{i=1}^n$:
\begin{equation}
	F_i = \phi \left( I_i, P_i \right).
\end{equation}
% where $ n $ is the batch size, $ I_i $ represents the $ i $-th image, $ E_i $ is the corresponding exemplar, and $ F_i $ denotes the feature produced by the encoder.

\vspace{-10pt}
\paragraph{Entropy Calculation.} 
Entropy, which measures feature complexity, generally increases with the amount of information in an image and is positively correlated with the number of categories. It is computed as follows:
\begin{equation}
\resizebox{0.9\columnwidth}{!}{$
	E_i = -\sum_{h=1}^H \sum_{w=1}^W \sum_{c=1}^C \left[ F_i \log F_i + (1 - F_i) \log (1 - F_i) \right], $}
\end{equation}
where $E_i$ represents the entropy across the feature dimensions: $H$ for height, $W$ for width, and $C$ for channels.

\vspace{-10pt}
\paragraph{Offset Calculation.} 
The feature offset measures the spatial deviation of activations from the image's geometric center, indicating the distribution of features. Offset is positively correlated with category count; objects in single-category images are more centrally located, with centralization decreasing as category diversity increases:
\begin{equation}
	O_i = \left| \frac{\sum_{h=1}^H \sum_{w=1}^W w F_i}{\sum_{h=1}^H \sum_{w=1}^W F_i} - \frac{W}{2} \right|,
\end{equation}
where $O_i$ reflects the x-axis deviation of the centroid from the image center for each feature map $F_i$.

% \textbf{Feature Certainty}
% Certainty $ C $ evaluates the reliability of feature activations by distinguishing between robust and weak signal strengths. This measure is crucial for interpreting the significance and reliability of features on a probabilistic scale:
% \begin{equation}
% \bar{X}_i = \frac{\mu(X_i) - \min(X_i)}{\max(X_i) - \min(X_i)}
% \end{equation}
% \begin{equation}
% C_i = 1.0 - \mathrm{sigmoid}(\bar{X}_i)
% \end{equation}
% Where $ \bar{X}_i $ is the normalized mean value of the feature activations for the $ i $-th feature map, scaling the mean activation values to a [0, 1] range. $ \mu(X_i) $ denotes the mean of $ X_i $, and the certainty $C_i$ is computed using a sigmoid function to convert the normalized mean into a probability-like measure, where values closer to 1 indicate higher certainty or confidence in the activations. This metric highlights regions of the image where the model detects features with high confidence, facilitating more reliable interpretations and decisions based on the feature maps.

\vspace{-10pt}
\paragraph{Certainty Calculation.} 
Certainty measures the confidence in predictions and is negatively correlated with category count. The model exhibits lower confidence in areas with unspecified categories. Certainty $C_i$ is derived by applying a sigmoid function to the normalized mean magnitude of activations $\mu_i$, as follows:
\begin{equation}
	S_i = \sum_{h=1}^H \sum_{w=1}^W \left| F_{i,h,w} \right|, \quad \mu_i = \frac{S_i}{H \times W},
\end{equation}
where $S_i$ is the sum of absolute activations in the $i$-th feature map, and $\mu_i$ represents the normalized mean magnitude of activations. Finally, feature certainty $C_i$ is computed by applying a sigmoid function to $\mu_i$, followed by inversion:
\begin{equation}
	C_i = 1 - \mathrm{sigmoid} \left( \mu_i \right).
\end{equation}

\subsection{Dirichlet Mixture}
Although entropy, offset, and certainty correlate with category count, these relations are not absolute. To achieve a more reliable evaluation, we combine these attributes into a composite weight using the Dirichlet distribution. Given attributes $E_i$ (entropy), $O_i$ (offset), and $C_i$ (certainty), the Dirichlet distribution is parameterized by a concentration vector $\alpha = [\alpha_1, \alpha_2, \alpha_3]$:
\begin{equation}
	p \left( w \mid \alpha \right) = \frac{1}{B(\alpha)} \prod_{k=1}^3 w_k^{\alpha_k - 1},
\end{equation}
where $w = [w_1, w_2, w_3]$ is the weight vector with $\sum_{k=1}^3 w_k = 1$, and $B(\alpha)$ is the multivariate beta function:
\begin{equation}
	B(\alpha) = \frac{\prod_{k=1}^3 \Gamma \left( \alpha_k \right)}{\Gamma \left( \sum_{k=1}^3 \alpha_k \right)}.
\end{equation}

Sampling from the Dirichlet distribution:
\begin{equation}
	w \sim \mathrm{Dirichlet} (\alpha),
\end{equation}
where the sampled weight vector $w$ combines the attributes into a weighted representation:
\begin{equation}
	U_\mathcal{C} = \sum_{k=1}^3 w_k U_k,
\end{equation}
where $U_1 = E_i$, $U_2 = O_i$, and $U_3 = C_i$. The final $U_\mathcal{C}$ represents the combined certainty of the three attributes.

% This method ensures that the combined features are adaptively weighted, reflecting the relative importance of entropy, centroid shift, and certainty as determined by the Dirichlet distribution.
% To achieve a more reliable evaluation, we integrate these attributes using Dempster-Shafer evidence theory. The combined evidence set $W_i = \{\{b_c^i\}_{c = 1}^v, u^i\}$ is calculated from the attributes entropy ($E$), offset ($O$), and certainty ($C$) as follows:
% \begin{equation}
% W = DS(E, O, C)
% \end{equation}

% The final combined uncertainty for the batch is derived from these calculations, represented as $ W = \{w_i\}_{i = 0}^n $, providing a comprehensive uncertainty assessment for the entire batch of images.

\subsection{Error-Sensitive Focal-MSE}
\label{sec 3.2}

To enhance model sensitivity to target regions and improve accuracy in object counting, we introduce {Focal-MSE}, a novel loss function that modifies traditional MSE loss. The predicted density map $M^p$ and ground truth map $M^g_i$ are treated as matrices, where each entry in $M^g_i$ indicates object presence, and $M^p$ represents the predicted probability. {Focal-MSE} focuses on sparsely populated and error-prone areas to ensure precise counting. The error-sensitive (ES) loss is computed as:
\begin{equation}
\begin{aligned}
	\mathcal{L}^\mathrm{ES}_i \left( M^p_i, M^g_i \right) = -\sum_{h=1}^H \sum_{w=1}^W \Big[ M^g_i \log \left( M^p_i \right) \\ + \left(1 - M^g_i \right) \log \left( 1 - M^p_i \right) \Big].
\end{aligned}
\end{equation}

The {Focal-MSE} loss, combining MSE loss with ES loss, is defined as:
\begin{equation}
	\mathcal{L}^\mathrm{FMSE}_i \left(M^p_i, M^g_i \right) = \sum_{h = 1}^H \sum_{w = 1}^W \left(M^p_i - M^g_i \right)^2 + \mathcal{L}^\mathrm{ES}_i.
\end{equation}

{Focal-MSE} enhances category information, although it may not yield precise density maps for object counting. MSE loss is therefore introduced in the second phase to ensure accuracy:
\begin{equation}
\mathcal{L}^\mathrm{MSE}_i \left(M^p_i, M^g_i \right) = -\sum_{h = 1}^H \sum_{w = 1}^W 
 \left(M^p_i - M^g_i \right)^2.
\end{equation}

\vspace{-10pt}
\paragraph{Dual-phase Curriculum Loss.}
FocalCount employs a curriculum learning scheme for enhanced supervision and sensitivity to non-specified category errors, using a dual-phase loss. Focal-MSE is effective in early training stages with ambiguous information, while MSE loss is prioritized later to refine predictions. Assuming $t$ epochs for the first phase, the dual-phase loss function is:
\begin{equation}
	\mathcal{L}^D_i = \left\{
	\begin{array}{ll}
	\mathcal{L}^\mathrm{FMSE}_i \left(M^p_i, M^g_i \right), & \mathrm{if \quad epoch} < t, \\
	\mathcal{L}^\mathrm{MSE}_i \left(M^p_i, M^g_i \right), & \mathrm{if \quad epoch} > t.
	\end{array}
	\right.
\end{equation}

\vspace{-10pt}
\paragraph{Overall Loss.}

To counteract category imbalance within counting datasets, which biases models toward indiscriminate counting, we propose a dual-phase loss for regions combined with uncertainty-based category count evaluation. The overall loss of FocalCount integrates both strategies:
\begin{equation}
	\mathcal{L}_\mathrm{all} = \frac{1}{n} \sum_{i = 1}^n \left[ U_\mathcal{C} \mathcal{L}^D_i \left(M^p_i, M^g_i \right) \right], 
\end{equation}

\section{Experimental Result}

\subsection{Datasets}

\textsc{FSC-147}~\cite{hobley2022learning} dataset is a class-agnostic counting dataset consisting of 6,135 images across 147 classes. It features non-overlapping subsets and provides dot annotations for zero-shot counting.

\textsc{CARPK}~\cite{hsieh2017drone} dataset includes 89,777 cars in 1,448 parking lot images, serving as an ideal benchmark for testing cross-dataset transferability.

\textsc{ShanghaiTech}~\cite{zhang2016single} dataset is a crowd counting dataset with two parts: Part A (\textsc{SHA}) containing 482 images and Part B (\textsc{SHB}) containing 716 images. Each part contains 400 training images, with the remainder used for testing. Cross-part evaluations are challenging due to different data collection methods.

% \textsc{FSC-147}~\cite{hobley2022learning} is a dataset specifically designed for class-agnostic counting, containing 6, 135 images across 147 classes. It is unique in its use of non-overlapping class subsets and provides class labels and dot annotations for zero-shot counting using textual prompts. 

% \textsc{CARPK}~\cite{hsieh2017drone} offers a bird's-eye view of 89, 777 cars in 1, 448 parking lot images, making it ideal for testing the cross-dataset transferability and adaptability of counting methods. 

% \textsc{ShanghaiTech}~\cite{zhang2016single} is a widely used crowd counting dataset that includes Part A (482 images) and Part B (716 images). Each part has 400 training images, with the remainder used for testing. The different collection methods add complexity to cross-part evaluations. In the text, Part A is referred to as SHA, and Part B as SHB.

\subsection{Evaluation Metrics.}
We use mean absolute error (MAE) and root mean square error (RMSE) as evaluation metrics. MAE measures accuracy, while RMSE assesses robustness, following previous class-agnostic object counting methods~\cite{nguyen2022few}.

\subsection{Implementation Details} 
We evaluate our loss function on \textsc{FSC-147} under both zero-shot and few-shot settings. Training uses a batch size of 32 and the AdamW~\cite{loshchilov2017decoupled} optimizer, starting with a learning rate of 1e-4, decaying by a factor of 0.33 after 100 epochs. The training lasts for 300 epochs, with the first 50 epochs serving as the initial stage. Zero-shot training follows the CLIP-Count~\cite{jiang2023clip} baseline, while few-shot training follows the CACViT~\cite{WangX0024} baseline, with a weight decay of 0.05 and 10 warm-up epochs. All experiments, including ablation studies, are conducted on an NVIDIA RTX L40 GPU.

\begin{table*}
	\centering
	\footnotesize
	\setlength{\tabcolsep}{11pt}
	\begin{tabular}{clcccccccc}
	\toprule[1.2pt]
	\multirow{2}[2]{*}{Scheme} & \multirow{2}[2]{*}{Method} & \multirow{2}[2]{*}{Venue} & \multirow{2}[2]{*}{Shot} & \multicolumn{2}{c}{Val Set} & \multicolumn{2}{c}{Test Set} & \multicolumn{2}{c}{Avg} \\
	\cmidrule(lr){5-6} \cmidrule(lr){7-8} \cmidrule(lr){9-10}
	 & & & & MAE & RMSE & MAE & RMSE & MAE & RMSE \\
	\midrule
	\multirow{4}{*}{Reference-less}
	 & FamNet~\cite{ranjan2021learning} & CVPR'21 & 0 & 32.15 & 98.75 & 32.27 & 131.46 & 32.21 & 115.11 \\
	 & RCC~\cite{hobley2022learning} & CVPR'22 & 0 & 17.49 & 58.81 & 17.12 & 104.53 & 17.31 & 81.67 \\
	 & CounTR~\cite{liu2022countr} & BMVC'22 & 0 & 18.07 & 71.84 & {14.71} & 106.87 & {16.39} & 89.36 \\
	 & LOCA~\cite{djukic2023low} & ICCV'23 & 0 & {17.43} & {54.96} & 16.22 & {103.96} & 16.83 & {79.46} \\
	\midrule
	\multirow{8}{*}{Few-shot}
	 & FamNet~\cite{ranjan2021learning} & CVPR'21 & 1 & 26.05 & 77.01 & 26.76 & 110.95 & 26.41 & 93.98 \\
	\cmidrule{2-10}
	 & FamNet~\cite{ranjan2021learning} & CVPR'21 & 3 & 24.32 & 70.94 & 22.56 & 101.54 & 23.44 & 86.24 \\
	 & CFOCNet~\cite{yang2021class} & WACV'21 & 3 & 21.19 & 61.41 & 22.10 & 112.71 & 21.65 & 87.06 \\
	 & CounTR~\cite{liu2022countr} & BMVC'22 & 3 & 13.13 & 49.83 & 11.95 & 91.23 & 12.54 & 70.53 \\
	 & PseCo~\cite{huang2023point} & CVPR'23 & 3 & 15.31 & 68.34 & 13.05 & 112.86 & 14.18 & 90.60 \\
	 & LOCA~\cite{djukic2023low} & ICCV'23 & 3 & \textbf{10.24} & \textbf{32.56} & 10.97 & \underline{56.97} & \underline{10.61} & \underline{44.77} \\
	 & CACViT~\cite{WangX0024}$^\dag$ & AAAI'24 & 3 & 11.41 & \underline{39.75} & \underline{10.47} & 66.24 & 10.94 & 52.99 \\
	 & CACViT + Ours & & 3 & \underline{11.07} & 41.79 & \textbf{10.13} & \textbf{43.85} & \textbf{10.60} & \textbf{42.82} \\
	\midrule
	\multirow{6}{*}{Zero-shot} 
	 & ZSC~\cite{xu2023zero} & CVPR'23 & 0 & 26.93 & 88.63 & 22.09 & {115.17} & 24.51 & 101.90 \\
 	 & VA-Count~\cite{zhu2024zero} & ECCV'24 & 0 & 17.87 & {73.22} & 17.88 & 129.31 & 17.87 & 101.26 \\	 & CounTX~\cite{amini2023open}$^\dag$ & BMVC'23 & 0 & 18.32 & 63.21 & 18.47 & 106.15 & 18.39 & 84.68 \\
	 & CounTX + Ours & & 0 & \textbf{17.19} & 62.96 & {17.63} & 110.64 & \textbf{17.41} & 86.80 \\ 
	 \cmidrule{2-10}	 
 	 & CLIP-Count~\cite{jiang2023clip}$^\dag$ & ACM MM'23 & 0 & {18.79} & \underline{61.18} & \underline{17.78} & \underline{106.62} & 18.29 & \underline{83.90} \\
	 & FocalCount (Ours) & & 0 & \underline{18.57} & \textbf{61.02} & \textbf{17.60} & \textbf{102.74} & \underline{18.08} & \textbf{81.88} \\
	\bottomrule[1.2pt]
	\end{tabular}
	\caption{\textbf{Quantitative Comparison of the FocalCount Method with State-of-the-Art Methods on \textsc{FSC-147}.} $\dag$ indicates models using the same backbone. The best and second-best results are highlighted in \textbf{bold} and \underline{underlined}, respectively. {Avg} represents the average performance across the test and validation sets.}
	\label{tab:ExpSOTA}
\end{table*}

\subsection{Comparison with State-of-the-Art Methods}
% To evaluate our method's performance, we benchmark it on \textsc{FSC-147} against a variety of state-of-the-art few-shot and zero-shot counting methods. We also extend our evaluations through transfer experiments on \textsc{CARPK} and \textsc{ShanghaiTech}, comparing our method with class-specific counting models.

\paragraph{Quantitative Results on \textsc{FSC-147}.}
\cref{tab:ExpSOTA} showcases FocalCount's performance against state-of-the-art methods on \textsc{FSC-147}, particularly excelling in both few-shot and zero-shot settings on unseen datasets. In few-shot scenarios, while validation performance matches the baseline, incorporating Focal-MSE into CounTX slightly boosts results and FocalCount achieves a 22-point reduction in test RMSE, enhancing its generalizability. This suggests a better handling of varied scenarios and reduced overfitting.
In zero-shot settings, FocalCount outperforms the baseline on both validation and test sets, with a 4-point RMSE reduction, indicating fewer large errors and improved prediction stability. Although its MAE is slightly higher than VA-Count~\cite{zhu2024zero}, the substantial 26-point drop in test RMSE highlights FocalCount's robustness and superior generalization from learned representations to new data.
The integration of Focal-MSE, though modest, underscores the potential for refining loss functions to improve counting accuracy, suggesting that even minor model adjustments can lead to significant performance enhancements.

\begin{table}
	\centering
	\setlength{\tabcolsep}{7pt}
	\footnotesize
	\begin{tabular}{lcccc}
	\toprule[1.2pt]
	\multirow{2}[2]{*}{Method} & \multirow{2}[2]{*}{Venue} & \multirow{2}[2]{*}{Shot} & \multicolumn{2}{c}{FSC $\to$ \textsc{CARPK}} \\ 
	\cmidrule(lr){4-5} 
	& & & MAE & RMSE \\
	\midrule 	
	FamNet~\cite{ranjan2021learning} & CVPR'21 & 3 & 28.84 & 44.47 \\
	BMNet~\cite{shi2022represent} & CVPR'22 & 3 & 14.41 & 24.60 \\
	BMNet+~\cite{shi2022represent} & CVPR'22 & 3 & {10.44} & {13.77} \\
 	\midrule 	
	RCC~\cite{hobley2022learning} & CVPR'22 & 0 & 21.38 & 26.61 \\
	CLIP-Count~\cite{jiang2023clip} & ACM MM'23 & 0 & 13.59 & 18.30 \\
	FocalCount (Ours) & & 0 & \textbf{12.74} & \textbf{15.33} \\
	\bottomrule[1.2pt]
	\end{tabular}
	\caption{\textbf{Quantitative Comparison of FocalCount with State-of-the-Art Methods on \textsc{CARPK}.}}
	\label{tab2}
\end{table}

\vspace{-10pt}
\paragraph{Generalizability Evaluation on \textsc{CARPK}.}

To evaluate the cross-dataset generalizability of FocalCount, we train the model on \textsc{FSC-147} and test it on \textsc{CARPK} without fine-tuning, following the protocol of previous class-agnostic counting approaches. As shown in \cref{tab2}, our method nearly matches the performance of few-shot methods on \textsc{CARPK}, achieving an MAE of 12.74 and an RMSE of 15.33. It outperforms other zero-shot methods, demonstrating robust generalizability.

\begin{table}
	\centering
	\setlength{\tabcolsep}{5pt}
	\footnotesize
	\begin{tabular}{clcccc}
	\toprule[1.2pt]
	\multirow{2}[2]{*}{Type} & \multirow{2}[2]{*}{Method} & \multicolumn{2}{c}{\textsc{SHB}} & \multicolumn{2}{c}{\textsc{SHA}} \\
	\cmidrule(lr){3-4} \cmidrule(lr){5-6}
	& & MAE & RMSE & MAE & RMSE \\
	\midrule
	\multirow{2}{*}{Specific} & MCNN~\cite{zhang2016single} & 85.20 & 142.30 & 221.40 & 357.80 \\
	& CrowdCLIP~\cite{liang2023crowdclip} & 69.60 & 80.70 & 217.00 & 322.70 \\
	\midrule
	\multirow{3}{*}{Generic} & RCC~\cite{hobley2022learning} & 66.60 & 104.80 & 240.10 & 366.90 \\
	& CLIP-Count~\cite{jiang2023clip} & 47.92 & 80.48 & 197.47 & 319.75 \\
	& FocalCount (Ours) & \textbf{45.87} & \textbf{77.68} & \textbf{185.90} & \textbf{291.72} \\
	\bottomrule[1.2pt]
	\end{tabular}
	\caption{\textbf{Cross-Dataset Evaluation on the \textsc{ShanghaiTech} Crowd Counting Dataset.} Generic models are trained on \textsc{FSC-147}, while specific models are trained on \textsc{SHA}.}
	\label{tab:3}
\end{table}

\vspace{-10pt}
\paragraph{Generalizability Evaluation on \textsc{ShanghaiTech}. }

To assess transferability and generalizability, we evaluate FocalCount on \textsc{ShanghaiTech}, following established protocols. As shown in \cref{tab:3}, our model, trained exclusively on \textsc{FSC-147}, outperforms both methods transferred from crowd datasets and other generic models trained on \textsc{FSC-147}. It achieves lower error rates and demonstrates superior generalizability, indicating its robustness in handling highly variable crowd datasets.

\begin{table}
	\centering
	\setlength{\tabcolsep}{3pt}
	\footnotesize
	\begin{tabular}{ccccccccc}
	\toprule[1.2pt]
	\multirow{2}[2]{*}{$\mathcal{L}^\mathrm{MSE}$} & \multirow{2}[2]{*}{$\mathcal{L}^\mathrm{ES}$} & \multirow{2}[2]{*}{$U_C$} & \multicolumn{2}{c}{Val Set} & \multicolumn{2}{c}{Test Set} & \multicolumn{2}{c}{Avg Set} \\
 	\cmidrule(lr){4-5} \cmidrule(lr){6-7} \cmidrule(lr){8-9}
	 & & & MAE & RMSE & MAE & RMSE & MAE & RMSE \\
	\midrule
	\CIRCLE & \Circle & \Circle & 18.78 & 62.23 & 18.29 & 105.74 & 18.54 & 83.98 \\
	\Circle & \CIRCLE & \Circle & \textbf{18.50} & 61.63 & 18.42 & {105.54} & 18.46 & 83.58 \\
	\Circle & \CIRCLE & \CIRCLE & 18.52 & \textbf{58.74} & 17.95 & 106.95 & 18.24 & 82.85 \\
	\CIRCLE & \CIRCLE & \CIRCLE & 18.57 & 61.02 & \textbf{17.60} & \textbf{102.74} & \textbf{18.08} & \textbf{81.88} \\
	\bottomrule[1.2pt]
	\end{tabular}
 	\caption{\textbf{Ablation Study on the Contribution of Each Component to the Final Results on \textsc{FSC-147}.} $U_C$ denotes integrated uncertainty. Focal-MSE is applied during the initial training phase, while both MSE loss and $U_C$ are utilized throughout the entire training process.}
	\label{tab:abalation_factors}
\end{table}

\subsection{Ablation Study}

% \subsubsection{Quantitative Results on \textsc{ShanghaiTech} after Transfer.}

\paragraph{Ablation Study on Various Factors.}

% \subsubsection{Ablation of Different Feature Attributes}

We conducted an ablation study to evaluate the contribution of each component in FocalCount, including Focal-MSE, MSE loss, ES loss, and multi-feature attribute uncertainty. \cref{tab:abalation_factors} presents the results. Using MSE loss alone results in an MAE of 18.78 (first row), while using ES loss alone improves the MAE to 18.50 (second row). Adding uncertainty with ES loss further reduces the MAE to 18.24. Optimal performance is achieved when all components are integrated, showing that Focal-MSE effectively improves accuracy by addressing category-specific errors and data imbalance.

\begin{table}
	\centering
	\footnotesize
	\setlength{\tabcolsep}{6pt}
	\begin{tabular}{ccccccc}
	\toprule[1.2pt]
	\multirow{2}[2]{*}{Epoch} & \multicolumn{2}{c}{Val Set} & \multicolumn{2}{c}{Test Set} & \multicolumn{2}{c}{Avg Set} \\
 	\cmidrule(lr){2-3} \cmidrule(lr){4-5} \cmidrule(lr){6-7}
	 & MAE & RMSE & MAE & RMSE & MAE & RMSE \\
	\midrule
	10 & 19.35 & 65.83 & 18.94 & 107.61 & 19.14 & 86.72 \\ 
	20 & \textbf{18.57} & \textbf{61.02} & 17.60 & 102.74 & 18.08 & \textbf{81.88} \\ 
	30 & 18.58 & 63.96 & \textbf{17.09} & \textbf{102.09} & \textbf{17.84} & 83.03 \\ 
	40 & 18.88 & 63.87 & 18.39 & 106.73 & 18.64 & 85.30 \\ 
	50 & 18.83 & 63.19 & 18.39 & 106.58 & 18.61 & 84.88 \\ 
	\bottomrule[1.2pt]
	\end{tabular} 
 	\caption{\textbf{Ablation Study on Different Pre-training Epochs.}}
	\label{tab1}
\end{table}

\vspace{-10pt}
\paragraph{Ablation Study on Pre-training Epochs.}

To determine the optimal number of pre-training epochs, we conducted experiments whose results are presented in \cref{tab1}. Both 20 and 30 epochs produced strong results, but 20 epochs were selected as the optimal choice due to their superior RMSE, indicating greater robustness. The model trained with only 10 epochs performed the weakest, underscoring the importance of sufficient pre-training.

\paragraph{Ablation Study on Computational Cost}
Tab.~\ref{tab3} presents the computational times per batch for various configurations involving the loss functions $\mathcal{L}^\mathrm{MSE}$, $\mathcal{L}^\mathrm{ES}$, and uncertainty $U_C$. The results demonstrate that these configurations introduce only a modest increase in computational time. The highest recorded time is 2.24 milliseconds, slightly above the base time of 70 microseconds. This minimal increase indicates that the integration of enhanced loss functions and uncertainty components does not significantly impact computational efficiency per batch. Consequently, our approach achieves improved model accuracy without incurring substantial computational overhead, ensuring efficient processing even with advanced algorithms.

\begin{table}[h]
	\centering
	\setlength{\tabcolsep}{10pt}
	\footnotesize
	\begin{tabular}{cccccc}
	\toprule
	\multirow{2}[2]{*}{$\mathcal{L}^\mathrm{MSE}$} & \multirow{2}[2]{*}{$\mathcal{L}^\mathrm{ES}$} & \multirow{2}[2]{*}{$U_C$} & \multirow{2}[2]{*}{Times} & \multicolumn{2}{c}{Avg Set} \\
 	\cmidrule(lr){5-6}
	 & & & & MAE & RMSE \\
	\midrule
	\CIRCLE & \Circle & \Circle & 70.00 us & 18.54 & 83.98 \\
	\Circle & \CIRCLE & \Circle & 69.00 us & 18.46 & 83.58 \\
	\CIRCLE & \CIRCLE & \Circle & 84.00 us& 18.06 & 83.45 \\
	\Circle & \CIRCLE & \CIRCLE & 1.55 ms & 18.24 & 82.85 \\
	\CIRCLE & \Circle & \CIRCLE & 1.50 ms & 18.10 & 85.53 \\
	\CIRCLE & \CIRCLE & \CIRCLE & 2.24 ms & \textbf{18.08} & \textbf{81.88} \\
	\bottomrule
	\end{tabular}
	\caption{Ablation study on the contribution of each component and time to the final results on \textsc{FSC-147}~\cite{hobley2022learning}. $U_C$ denotes the integrated uncertainty. Focal-MSE is applied during the initial training phase, while MSE loss and $U_C$ are employed throughout the entire process. Best results are highlighted in \textbf{bold}.}
	\label{tab3}
\end{table}
% \begin{table}
% 	\centering
% 	\small
% 	\caption{Ablation study on different uncertainties, where $U_1$, $U_2$, $U_3$, and $U_C$ represent feature entropy, feature offset, feature certainty, and combined uncertainty, respectively.}
% 	\setlength{\tabcolsep}{5pt}
% 	\begin{tabular}{lcccccccc}
% 	\toprule[1.2pt]
% 	\multirow{2}[2]{*}{Metric} & \multicolumn{2}{c}{Val Set} & \multicolumn{2}{c}{Test Set}& \multicolumn{2}{c}{Avg}\\
% 	\cmidrule(lr){2-3} \cmidrule(lr){4-5} 
% \cmidrule(lr){6-7}
% 	 & MAE & RMSE & MAE & RMSE & MAE & RMSE \\
% 	\midrule
% 	0 & 27.13 & 84.30 & 23.49 & 121.10 & 25.31 & 102.70\\ 
% 	1 & 26.49 & 85.46 & 22.02 & 116.43 & 24.26 & 100.95\\
% 	2 & 25.75 & 81.35 & 21.29 & 110.48 & 23.45 & 95.89\\
% 	3 & 25.03 & 77.87 & 21.87 & 120.24 & 23.45 & 99.06\\
% 	4 & 24.52 & 78.70 & 22.35 & 115.63 & 23.44 & 97.17\\
% 	5 & 24.09 & 79.39 & 21.16 & 109.69 & 22.62 & 94.54\\
% 0 & & & & & & \\
% 1 & & & & & & \\
% 2 & & & & & & \\
% 3 & & & & & & \\
% 4 & & & & & & \\
% 5 & & & & & & \\
% 	\bottomrule[1.2pt]
% 	\end{tabular} 
% 	\label{ta}
% \end{table}

\begin{table}
	\centering
	\footnotesize
 	\setlength{\tabcolsep}{6pt}
	\begin{tabular}{lcccccc}
	\toprule[1.2pt]
	\multirow{2}[2]{*}{Metric} & \multicolumn{2}{c}{Val Set} & \multicolumn{2}{c}{Test Set} & \multicolumn{2}{c}{Avg Set} \\
 	\cmidrule(lr){2-3} \cmidrule(lr){4-5} \cmidrule(lr){6-7}
	 & MAE & RMSE & MAE & RMSE & MAE & RMSE \\
	\midrule
	- & 18.78 & 62.23 & 18.29 & 105.74 & 18.53 & 83.99 \\
	$U_1$ & 18.65 & 64.14 & 17.84 & 106.29 & 18.24 & 85.22 \\
	$U_2$ & \textbf{18.22} & 64.08 & 17.77 & 106.99 & \textbf{17.99} & 85.53 \\
	$U_3$ & 18.50 & 63.37 & 17.67 & 105.72 & 18.09 & 84.55 \\
	$U_C$ & 18.57 & \textbf{61.02} & \textbf{17.60} & \textbf{102.74} & 18.08 & \textbf{81.88} \\ 
	\bottomrule[1.2pt]
	\end{tabular} 
 	\caption{\textbf{Ablation Study on Different Uncertainties.} $U_1$, $U_2$, $U_3$, and $U_C$ denote feature entropy, feature offset, feature certainty, and combined uncertainty, respectively.}
	\label{tab:abalation_metric}
\end{table}

\cref{tab:abalation_metric} compares different uncertainty methods. Each type of uncertainty leads to performance improvements, with the offset attribute $U_2$ providing the most significant enhancement. Combining all three uncertainties $U_C$ yields the best performance, demonstrating that integrating attribute uncertainty improves robustness and addresses data imbalance.

% \begin{figure}
% 	\centering
% 	\includegraphics[width = \linewidth]{figure/figure-10.pdf}
% 	\caption{\textbf{Impact of Number of Classes.} Performance evaluation on a training set of 1,000 images filtered with varying $U_C$ thresholds. From left to right, higher $U_C$ values indicate a greater likelihood of images containing multiple categories. Data is sourced from \textsc{FSC-147}.}
% 	\label{fig}
% \end{figure}
\begin{figure*}
	\centering
	\includegraphics[width = \linewidth]{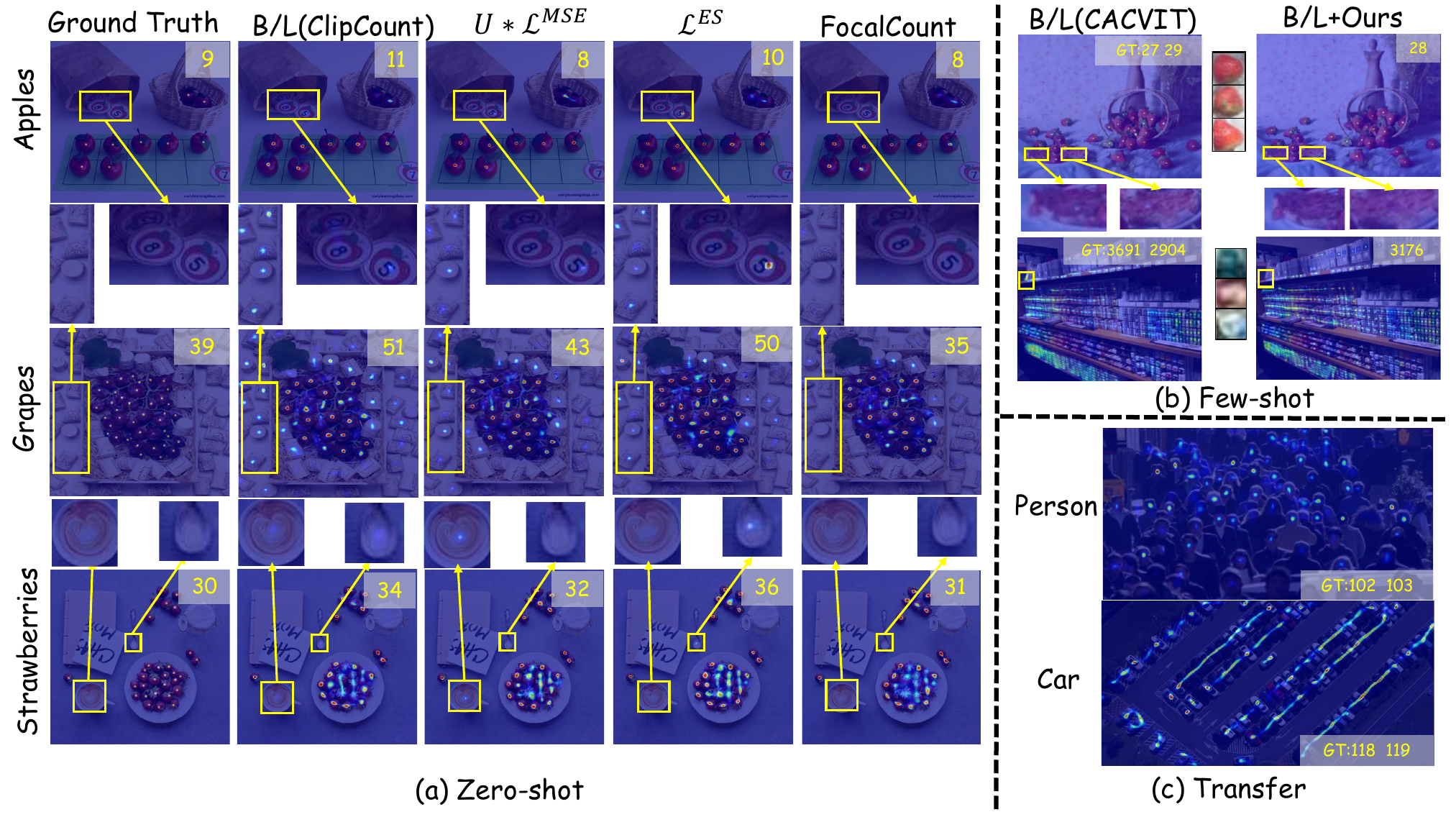}
	\caption{\textbf{Density Maps and Transfer Results in Zero-Shot and Few-Shot Settings.} 
	(a) Density maps for different components in the zero-shot setting, with yellow boxes highlighting error-prone regions. 
	(b) Comparison of density maps in the few-shot setting. 
	(c) Transfer results from \textsc{FSC-147} to \textsc{ShanghaiTech} and \textsc{CARPK}.}
	\label{fig:5}
\end{figure*}
\begin{figure}
	\centering
	\includegraphics[width = \linewidth]{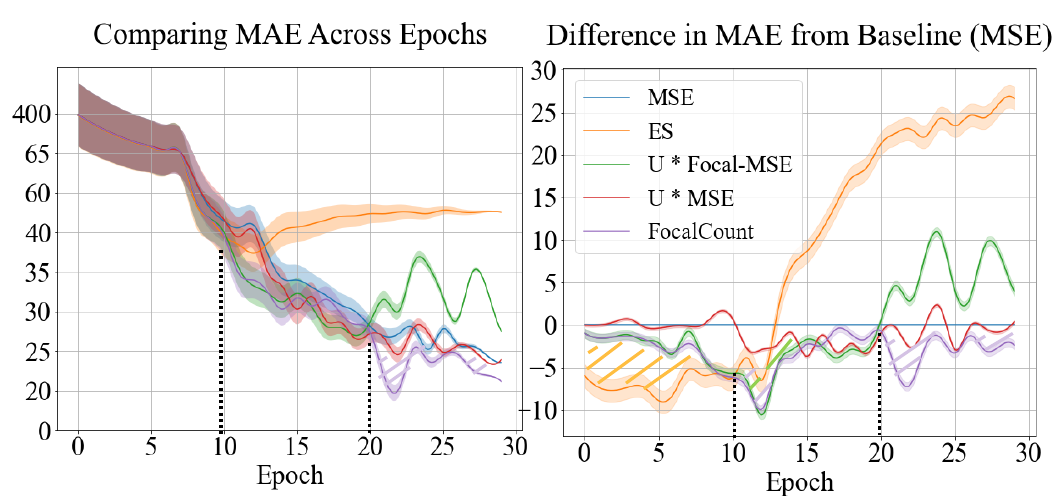}
	\caption{\textbf{Impact of Different Loss Functions on MAE during the First 30 Epochs.} The left panel shows the absolute impact, while the right panel compares the relative impact to the MSE loss. Y-axis intervals are adjusted for better visibility across a wide range of values.} 
	\label{fig:loss}
\end{figure}

% \subsection{Qualitative Analysis}
% \vspace{-10pt}
% \paragraph{Impact Analysis of Classes Number.}

% \cref{fig} illustrates the effect of $U_C$ on the performance of a training dataset with 1,000 images. $U_C$ quantifies the likelihood of an image containing 
%  multiple categories. In \textsc{FSC-147}, where most images contain a single category, increasing $U_C$ filters out more single-category images, enhancing dataset diversity and improving counting performance. The left plot shows a consistent decrease in MAE across the validation set, test set, and their average as $U_C$ increases. Similarly, the right plot demonstrates a reduction in RMSE, with minor fluctuations in the test set. These results highlight that higher $U_C$ values improve the model's generalization and counting accuracy.

\vspace{-10pt}
\paragraph{Impact Analysis of Loss Functions.}

\cref{fig:loss} shows the effect of different loss functions on model performance over the first 30 epochs, helping determine the optimal shift time in curriculum learning. During the first 10 epochs, ES loss (orange line) outperforms other loss functions, indicating the importance of early training in addressing error-prone regions. As training progresses, the advantage of ES loss diminishes, while MSE loss with uncertainty (orange line) consistently outperforms standard MSE loss in both MAE and RMSE. Within the first 20 epochs, Focal-MSE with uncertainty (red line) shows superior performance, surpassing the orange line. Ultimately, FocalCount (purple line) consistently outperforms standard MSE loss throughout training and surpasses all other loss functions.

% \subsection{Analysis of Feature and Class Count}

\vspace{-10pt}
\paragraph{Analysis of Specified Category Density Maps.}

\cref{fig:5}(a) displays density maps under various settings, demonstrating the effectiveness of our method in reducing errors in non-specified categories. The first row shows improved differentiation between apple slices and whole apples, outperforming traditional reweighting techniques. The second row shows a significant reduction in miscounts of sharp-edged candles, outperforming both the baseline and slight improvements over Focal-MSE. The third row, in a multi-category setting, illustrates effective differentiation between closely sized items like cups and spoons, outperforming both the baseline and partial Focal-MSE integration. \cref{fig:5}(b) and (c) further illustrate our method's efficacy in few-shot and transfer experiments, with reduced errors in challenging contexts. Our approach also demonstrates robustness in zero-shot and few-shot transfer, as shown in \cref{fig:5}(c), where our model successfully adapts to unseen datasets such as \textsc{ShanghaiTech} and \textsc{CARPK}.

% \subsection{Analysis of Comparative Method Performance}

\section{Conclusion}
This paper introduces {FocalCount}, a novel method for addressing data imbalances in class-agnostic object counting by leveraging diverse feature attributes to accurately estimate object category counts within images. These estimates are used as weighted factors to correct class-count disparities. We also propose {Focal-MSE}, an enhanced loss function that refines traditional MSE by incorporating larger error gradients, thereby increasing sensitivity to errors in unspecified categories. By integrating curriculum learning, {FocalCount} improves quantity supervision and further enhances the model's sensitivity to such errors.
Extensive experiments on three object counting datasets, including few-shot and zero-shot scenarios, validate the effectiveness of our approach. Notably, transfer tests on crowd and vehicle datasets demonstrate its robustness and generalizability. Future work will focus on advancing the model's capabilities in category-discovery counting, extending beyond counting specific target objects.

\section*{Acknowledgments}
This work was supported in part by the National Natural Science Foundation of China under Grant 62472332 and 62271361 and the Hubei Provincial Key Research and Development Program under Grant 2024BAB039.

% \newpage
{
 \small
 \bibliographystyle{ieeenat_fullname}
 \bibliography{FocalCount}
}

% \bibliography{main}
% \bibliographystyle{ieeenat_fullname}
% \bibliography{main}

\setcounter{page}{1}
\setcounter{section}{0}
\maketitlesupplementary

\section{Evaluation Metrics}
In accordance with previous class-agnostic object counting methods~\cite{nguyen2022few}, we utilize the Mean Absolute Error (MAE) and Root Mean Square Error (RMSE) as evaluation metrics. MAE quantifies the average magnitude of prediction errors, thereby reflecting the model's accuracy. In contrast, RMSE emphasizes larger errors, providing insights into the model's robustness. The formal definitions of these metrics are presented in \eqref{eq-mae} and \eqref{eq-rmse}, where $N$ denotes the number of images in the dataset, $C_i$ represents the predicted count of objects in the $i$-th image, and $C_i^\mathrm{GT}$ is the corresponding ground-truth count.
\begin{equation}
	\mathrm{MAE} = \frac{1}{N} \sum_{i=1}^N \left| C_i - C_i^\mathrm{GT} \right|,
	\label{eq-mae}
\end{equation}
\begin{equation}
	\mathrm{RMSE} = \sqrt{\frac{1}{N} \sum_{i=1}^N \left( C_i - C_i^\mathrm{GT} \right)^2}.
	\label{eq-rmse}
\end{equation}

\section{Ablation Study on Uncertainties}
To evaluate the effectiveness of our uncertainty selection and combination strategies, we present the results in \cref{tab2}. Instead of using $U_C$, we applied a weighted average method to combine the uncertainties. Among the individual uncertainty components, $U_2$ demonstrated the most effective performance. However, pairwise combinations did not yield improvements, likely due to interference between uncertainties. Although integrating all three uncertainties achieved competitive performance, it did not surpass the effectiveness of using a single uncertainty component, as interference persisted. In contrast, our approach leverages a Dirichlet distribution to combine uncertainties, achieving optimal results and outperforming methods that rely solely on individual uncertainty components. This underscores the efficacy and necessity of using a Dirichlet distribution for effectively managing combined uncertainties.
 
\begin{table}[h]
	\centering
	\setlength{\tabcolsep}{3.8pt}
	\footnotesize
	\begin{tabular}{cccccccccc}
	\toprule
	\multirow{2}[2]{*}{$U_1$} & \multirow{2}[2]{*}{$U_2$} & 
	\multirow{2}[2]{*}{$U_3$} & \multirow{2}[2]{*}{$U_C$} & \multicolumn{2}{c}{Val Set} & \multicolumn{2}{c}{Test Set} & \multicolumn{2}{c}{Avg Set} \\
 	\cmidrule(lr){5-6} \cmidrule(lr){7-8} \cmidrule(lr){9-10}
	& & & & MAE & RMSE & MAE & RMSE & MAE & RMSE \\
	\midrule
	\CIRCLE & \Circle & \Circle & \Circle & 18.65 & {64.14} & {17.84} & 106.29 & {18.24} & {85.22} \\
	\Circle & \CIRCLE & \Circle & \Circle & \textbf{18.22} & 64.08 & 17.77 & 106.99 & \textbf{17.99} & 85.53 \\
	\Circle & \Circle & \CIRCLE & \Circle & {18.50} & 63.37 & 17.67 & {105.72} & 18.09 & 84.55 \\
	\Circle & \CIRCLE & \CIRCLE & \Circle & 19.17 & 69.07 & 18.62 & 110.01 & 18.89 & 89.54 \\
	\CIRCLE & \Circle & \CIRCLE & \Circle & 18.99 & 67.39 & 18.51 & 114.98 & 18.75 & 91.18 \\
	\CIRCLE & \CIRCLE & \Circle & \Circle & 18.90 & 68.06 & 18.24 & 112.80 & 18.57 & 90.43 \\
	\CIRCLE & \CIRCLE & \CIRCLE & \Circle & 18.55 & 66.27 & 18.13 & 110.11 & 18.34 & 88.19 \\
	\CIRCLE & \CIRCLE & \CIRCLE & \CIRCLE & 18.57 & \textbf{61.02} & \textbf{17.60} & \textbf{102.74} & 18.08 & \textbf{81.88} \\
	\bottomrule
	\end{tabular}
	\caption{\textbf{Ablation Study on Different Uncertainties.} $U_1$, $U_2$, $U_3$, and $U_C$ denote feature entropy, feature offset, feature certainty, and combined uncertainty, respectively. The best results are highlighted in \textbf{bold}.}
	\label{tab2}
\end{table}

\section{Analysis of Density Maps}
\begin{figure}[h]
	\centering
	\includegraphics[width = \linewidth]{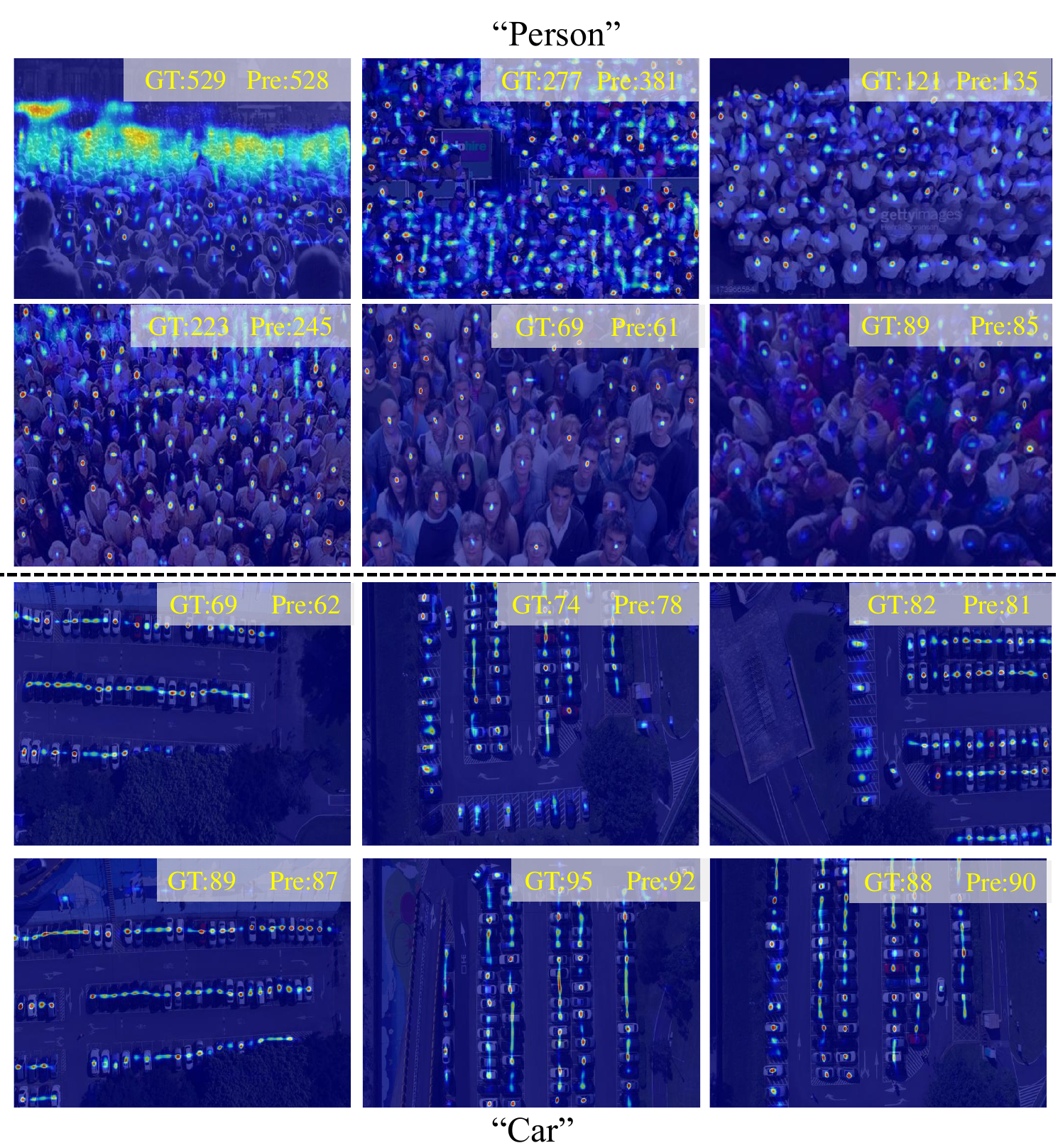}
	\caption{\textbf{Density Maps of Transfer to Human and Vehicle Categories.}}
	\label{fig2}
\end{figure}

\begin{figure*}[h]
	\centering
	\includegraphics[width = 0.75\textwidth]{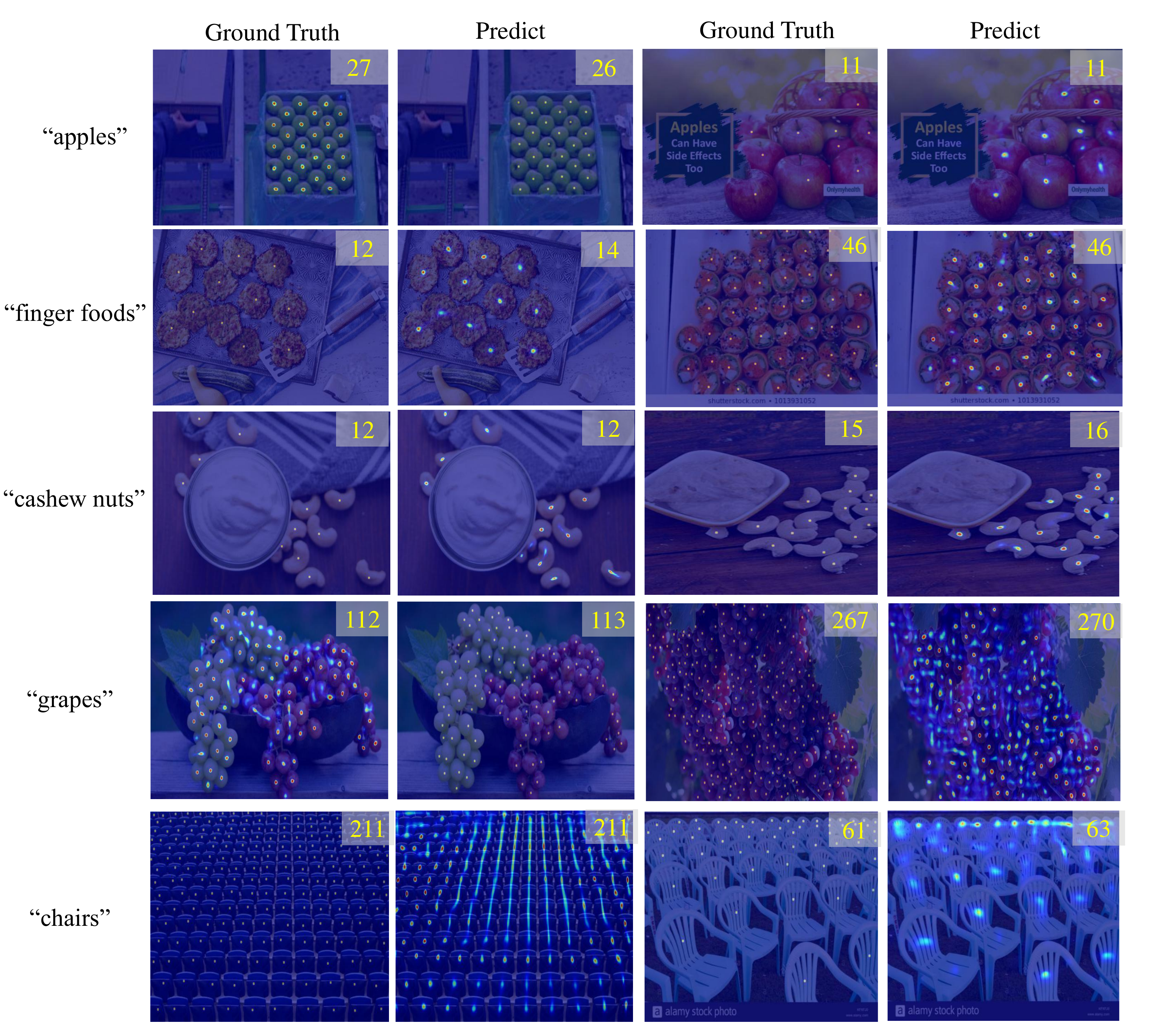}
	\caption{\textbf{Density Maps of Specified Categories in the Zero-Shot Setting.}
}
	\label{fig1}
\end{figure*}

\begin{figure*}[h]
	\centering
	\includegraphics[width = 0.75\textwidth]{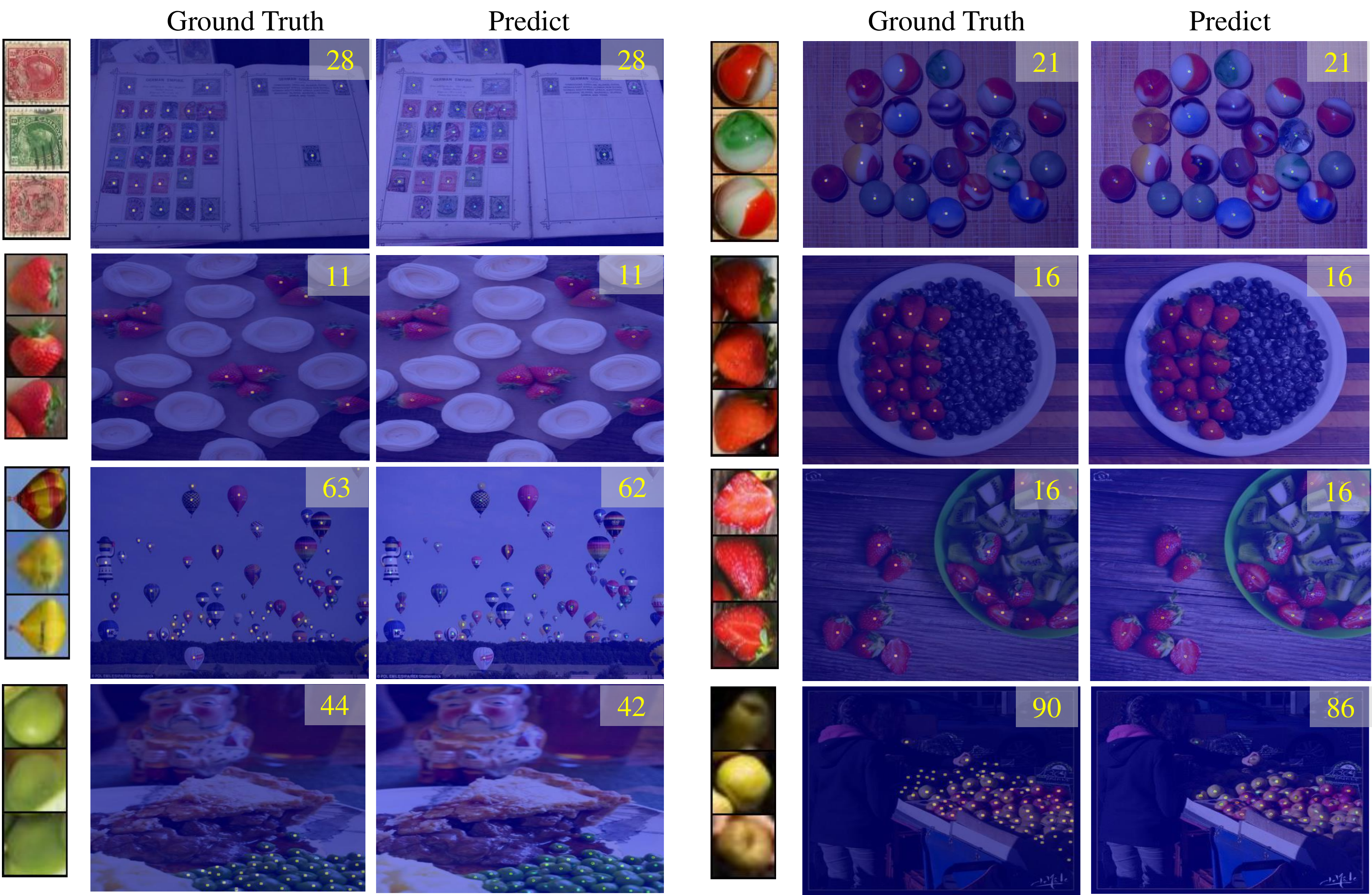}
	\caption{\textbf{Density Maps of Specified Categories in the Few-Shot Setting.}}
	\label{fig3}
\end{figure*}

To demonstrate the performance of our method across zero-shot, transfer, and few-shot scenarios, we present \cref{fig2},~\cref{fig1}, and~\cref{fig3}. These figures illustrate our model's ability to accurately distinguish between different object categories in specified density maps, highlighting its robustness and adaptability under various testing conditions.

\cref{fig2} shows density maps resulting from transferring our model directly from \textsc{FSC-147}~\cite{hobley2022learning} to \textsc{ShanghaiTech}~\cite{zhang2016single} and \textsc{CARPK}~\cite{hsieh2017drone} without additional training. Despite the challenges posed by dense crowd scenarios, our transfer experiments demonstrate the method's effectiveness, particularly in the vehicle dataset, which exhibited even fewer errors.

\cref{fig1} highlights the model's capability to distinguish and quantify different object categories in a zero-shot setting. This figure demonstrates the model's proficiency in handling objects with diverse densities and configurations by comparing predicted counts against ground truths across various categories, including apples, finger foods, cashew nuts, grapes, and chairs. The results indicate the model's potential for accurate object recognition and counting without prior category-specific training, a crucial capability for real-world applications where new object types frequently emerge.

\cref{fig3} showcases the model's performance in a few-shot setting, where it accurately identifies and counts different categories within images. This underscores the model's precision in distinguishing between various objects under limited training scenarios, ranging from everyday items like plates and fruits to more complex scenes involving balloons and marketplaces. The close alignment between predicted counts and ground truths across diverse scenes underscores the effectiveness of the few-shot learning approach in handling a variety of visual contexts with minimal prior data exposure.

\end{document}